\documentclass[fleqn,10pt]{wlscirep}

\usepackage{mdframed}

\usepackage{natbib}

\setcitestyle{numbers}
\title{Learning from data with structured missingness}

\author[1,2,*]{Robin Mitra}
\author[3, *]{Sarah F. McGough}
\author[1,4]{Tapabrata Chakraborti}
\author[1,5]{Chris Holmes}
\author[3]{Ryan Copping}
\author[6]{Niels Hagenbuch} 
\author[7]{Stefanie Biedermann}
\author[8]{Jack Noonan}
\author[2]{Brieuc Lehmann} 
\author[9]{Aditi Shenvi}
\author[10]{Xuan Vinh Doan}
\author[1,11]{David Leslie}
\author[1,12]{Ginestra Bianconi}
\author[1,13]{Ruben Sanchez-Garcia}
\author[1,14,15]{Alisha Davies}
\author[1, 16]{Maxine Mackintosh}
\author[17]{Eleni-Rosalina Andrinopoulou}
\author[1, 18]{Anahid Basiri}
\author[19,*]{Chris Harbron}
\author[1,13,20,*]{Ben D. MacArthur}

\affil[1]{The Alan Turing Institute, London, UK}
\affil[2]{Statistical Science, University College London, London, UK}
\affil[3]{Genentech, South San Francisco, CA, USA}
\affil[4]{Dept of Medical Physics \& Biomedical Engg. and UCL Cancer Institute, University College London, London, UK}
\affil[5]{Department of Statistics, University of Oxford, Oxford, UK}
\affil[6]{F.Hoffmann-La Roche AG, Basel Switzerland}
\affil[7]{School of Mathematics and Statistics, The Open University, Milton Keynes, UK}
\affil[8]{School of Mathematics, Cardiff University, Cardiff, UK}
\affil[9]{Department of Statistics, University of Warwick, Coventry, UK}
\affil[10]{Warwick Business School, University of Warwick 
Coventry, UK}
\affil[11]{The Digital Environment Research Institute, Queen Mary University of London, London, UK}
\affil[12]{School of Mathematical Sciences, Queen Mary University of London, London, UK}
\affil[13]{Mathematical Sciences, University of Southampton, Southampton, UK}
\affil[14]{Faculty of Health and Life Sciences, Swansea University, Swansea, UK}
\affil[15]{Public health Wales, Cardiff, UK}
\affil[16]{Genomics England, Dawson Hall, Charterhouse Square, London, UK}
\affil[17]{Department of Biostatistics and Department of Epidemiology, Erasmus MC, Rotterdam, Netherlands}
\affil[18]{School of Geographical \& Earth Sciences, University of Glasgow, UK}
\affil[19]{Roche Pharmaceuticals, Welwyn Garden City, UK}
\affil[20]{Faculty of Medicine, University of Southampton, Southampton, UK}

\affil[*]{Correspondence to: Robin Mitra (\href{ucakrmi5@ucl.ac.uk}{ucakrmi@ucl.ac.uk}), Sarah McGough (\href{mcgough.sarah@gene.com}{mcgough.sarah@gene.com}), Chris Harbron (\href{chris.harbron@roche.com}{chris.harbron@roche.com}) and Ben MacArthur (\href{bmacarthur@turing.ac.uk}{bmacarthur@turing.ac.uk})}

\begin{abstract}
Missing data are an unavoidable complication in many machine learning tasks. When data are `missing at random' there exist a range of tools and techniques to deal with the issue. However, as machine learning studies become more ambitious, and seek to learn from ever-larger volumes of heterogeneous data, an increasingly encountered problem arises in which missing values exhibit an association or structure, either explicitly or implicitly. Such `structured missingness' raises a range of challenges that have not yet been systematically addressed, and presents a fundamental hindrance to machine learning at scale. Here, we outline the current literature and propose a set of grand challenges in learning from data with structured missingness.   
\end{abstract}

\begin{document}

\maketitle

\section{Introduction}
Dealing with missing data is a longstanding problem in statistics and machine learning (ML) \cite{littlerubin2019, karlavs2020}. There are a wide range of ways to handle data that are `missing at random' \cite{rubin1976}, based upon well-established theory \cite{pigott2001, schafer2002, littlerubin2019, heitjan1991}. However, in many ML problems data may not be missing at random, but rather may exhibit some multivariate structure or pattern \cite{emmanuel2021survey}. These issues are particularly acute in areas that require extensive data fusion, multi-view learning, and/or data linkage that draw information from multiple sources and studies \cite{gao2020survey, yan2021deep, xu2013survey}. To advance in these areas we will need to solve the problem of assimilating and learning from data with highly `structured missingness' (SM). 

\begin{figure}
\centering
\includegraphics[width=\textwidth]{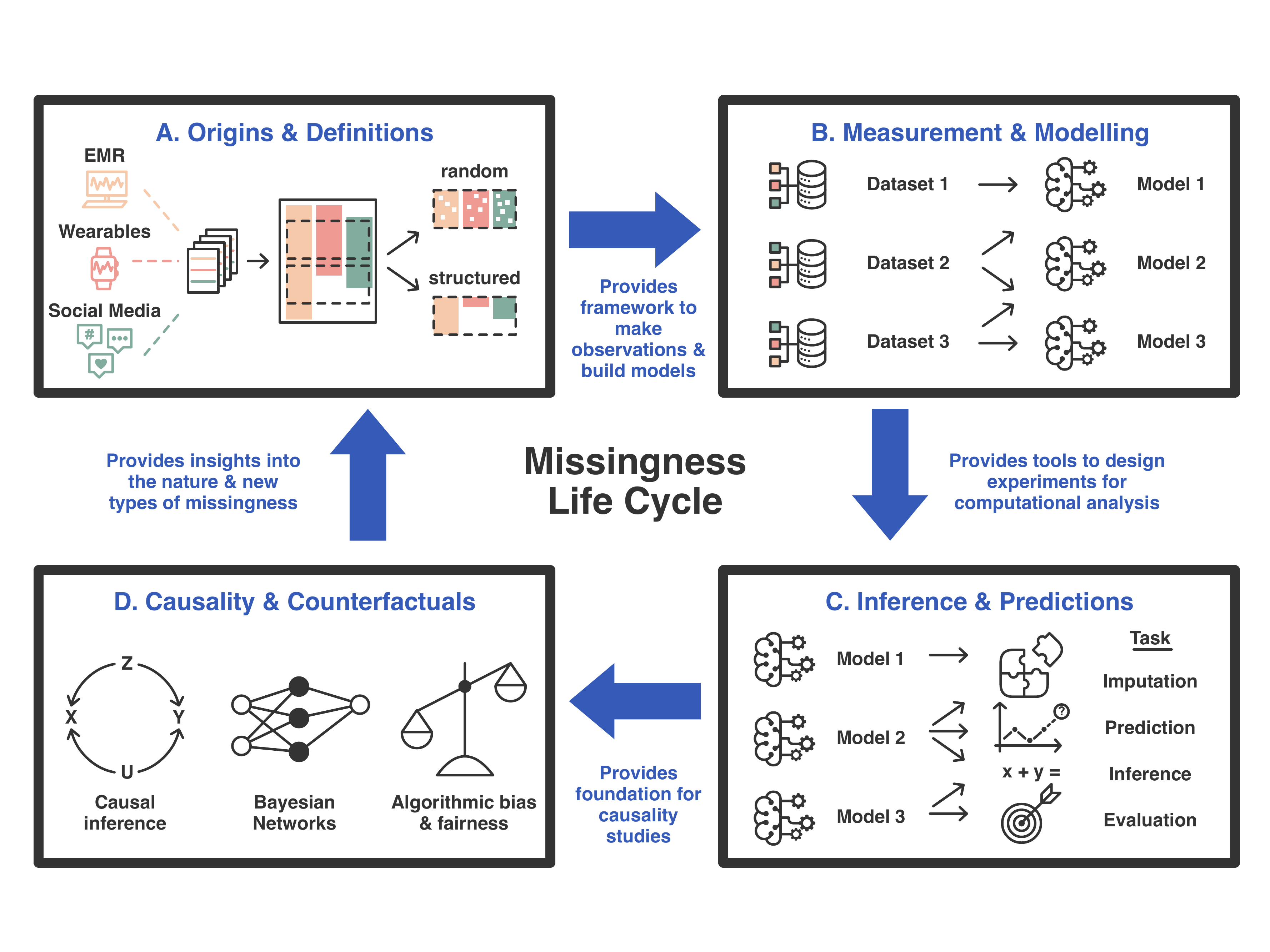}
\caption{\textbf{The data missingness life cycle.} The presence of SM affects all aspects of the data collection and analysis life cycle. (\textbf{A}) Data may be collected from numerous different sources -- in this illustration from electronic medical records, wearable devices and social media. When linked these data may give rise to both random and structured patterns of missingness. Understanding the interplay between data missing at random and SM requires new theory and tools, including new design tools to minimise the effects of SM on subsequent analysis. Challenges 1-3 relate to these issues. (\textbf{B}) SM affects our ability to construct models that can learn from the data appropriately and in an unbiased way. New tools to appropriately handle SM and new models that adapt to SM in the context of prediction or inference are needed. Challenges 4-5 relate to these issues. (\textbf{C}) Model efficacy will often (although not always) depend on our ability to impute missing readings accurately. New SM imputation tools are therefore needed, along with tools to benchmark and evaluate the performance of methods that model and deal with SM, including prediction and inference tasks. Challenges 4-7 address these issues. (\textbf{D}) In many scientific contexts the ultimate aim of any learning is to develop a better understanding of causality. The presence of SM can significantly compromise this effort. We require better tools to quantify and infer causal structures from data with SM, as well as methods to assess the extent to which missing data relates to systemic biases in data capture processes, which can, in turn, hamper the inference of causal mechanisms and lead to biased or unfair model outcomes. Challenges 8-9 relate to these issues. Taken together, addressing challenges at each stage of the missingness life cycle is vital to ensure appropriate analysis and insight generation from data with SM.}
\label{SMplot1}
\end{figure}

SM can occur for numerous reasons and so is a widely encountered issue. In \textbf{Box} \ref{B1} we outline some common routes to SM, that reflect the variety of ways in which SM can naturally arise in modern ML and modelling studies. Perhaps most pertinently for ML at scale, SM naturally arises when combining multi-modal datasets or when the data describe properties of a heterogeneous group of individuals with different characteristics. For instance, in a health context, SM arises when linking longitudinal clinical, genomic and imaging data \cite{topol2019high} or when approaching broad remit tasks, such as developing pan-cancer predictive models which incorporate diagnostic measurements for a range of different cancers and which are therefore only provided for subsets of patients \cite{silva2020pan}. More generally, SM commonly arises when combining information from multiple studies, each of which may vary in its design and measurement set and therefore only contain a sub-set of variables from the union of measurement modalities. In these situations, missing values may relate to the various different sampling methodologies used to collect the data, or reflect characteristics of the wider population of interest and so may impart useful information. Despite this possibility, the presence of SM typically significantly inhibits our ability to make effective use of data using current ML methods, and can severely impede all aspects of the analysis pipeline including building inferential models, designing predictive and classification algorithms, and producing informative visual data summaries. An overview of the various challenges associated with the data missingness life cycle is illustrated in \textbf{Fig}. \ref{SMplot1}. 

Established methods for dealing with missing data, such as (multiple) imputation \cite{rubin1996multiple}, do not usually take into account the structure of the missing data, and so may not deal with SM appropriately or effectively. As we move toward developing ML models that can learn from massive datasets at scale and perform numerous downstream tasks, such as foundational models \cite{bommasani2021opportunities}, and/or take federated approaches to learning \cite{kaissis2020secure, li2020federated}, these issues will become more acute. Indeed, it is not an exaggeration to say that issues of missingness are a primary hindrance to efficient learning at scale \cite{holmes2019}. However, despite the prevalence of this issue, SM has not yet been systematically studied and we lack both a theory for SM and the tools need to learn efficiently from data with SM. 

Where it does exist, the relevant literature tends to be sparsely scattered across different disciplines and typically focuses on re-purposing existing methodology to handle specific SM problems. For example, \cite{dong2019} consider adaptations to $k$-nearest neighbour methods to impute block missing data that arises when combining data from multiple high-throughput `omics' experiments. In a similar biomedical context, \cite{naito2021} draw on tools from multitask learning to impute blocks of missing genotypes. From a more theoretical perspective, \cite{audigier2018} consider adaptations to a multi-level model to multiply impute missing values when they arise systematically. Relatedly, \cite{kamphuis2018} propose a way to impute block missing values that extends classical multiple imputation methodology. More generally, the power of ML methods such as tree-based tools and deep learning to impute missing values and/or understand missingness structures is starting to be appreciated, although this literature is still embryonic \cite{che2018,wang2021,tierney2015}. Thus, while studies to date have partially addressed the problems of SM in particular settings, the methods proposed are typically bespoke, and while they make important advances toward understanding SM, their generalisability remains unclear. 

We believe that to advance ML at scale, across different areas of application and deployment, a more general approach to SM is needed that uses methods from across the computational sciences to better understand SM, how it affects subsequent learning and how it can be efficiently handled -- and even, since patterns of SM can convey important information, conditions under which it might be useful. For this reason, the Alan Turing Institute hosted a series of workshops to convene a multidisciplinary community of experts to identify the outstanding problems in SM, and sketch a road map for their solution. This paper is the fruit of those meetings, written by members of the community.

\textcolor{black}{The article is organised as follows. In section \ref{S1} we provide an introduction to SM, including a review of some fundamental concepts in missing data that are needed to understand SM, a description of common sources of SM, and an associated taxonomy of SM to conceptualise the issues involved. Introductory concepts are illustrated with examples, including a complex clinico-genomic data set that exemplifies well some of the challenges of SM \citep{singal2017cgdb}. These examples highlight how easily SM can occur, why the challenges of SM cannot be addressed with standard methodology, and the potential impact of SM on subsequent learning. In section \ref{S3} we outline a set of grand SM challenges, which if resolved will substantially advance the field, and place us in a stronger position to develop the next generation of ML methods that are able to learn efficiently from data at scale. We finish in section \ref{S4} with some forward-looking concluding remarks.}

\begin{mdframed}
\subsection*{Box 1: Routes to structured missingness} \label{B1}
Structured missingness (SM) is a broad term that refers to non-random multivariate patterns of missingness in a dataset, and so can arise in many ways, each with its own characteristics and challenges. Nevertheless, there are some common processes that give rise to SM in ML contexts. Here, we outline five.

\emph{Multi-modal linkage:} First, SM naturally arises whenever data captured from different sources, processes or experiments are linked together. Such data linkage is becoming increasingly common, and linked data are used to train some of the most powerful modern ML models. In this case, each unit in the data may only be measured for a subset of the available measurement modalities, but unobserved for the others. Box \ref{B2} gives an illustrative example of the challenges that this kind of SM poses in a large biomedical dataset \cite{singal2017cgdb, birnbaum2020modelassisted}.

\emph{Multi-scale linkage:} Second, SM naturally arises whenever measurements on different spatial or temporal scales are amalgamated. For instance, in geospatial analysis different sensors may measure different aspects of the climate. Some of these measurements -- such as sea surface temperature which can be relatively easily obtained from satellite thermal infrared sensors \cite{alerskans2020construction} -- may be well resolved in space and time, while others, such as precipitation, are harder to obtain and may be more coarsely resolved \cite{katiraie2013evaluation}. When considered collectively, the resulting data has characteristic patterns of missingness that relate to disparities in the precision and complexity of different sensor technologies.

\emph{Batch failure:} Third, SM may arise due to failures or discrepancies in data capture processes, such as sensor malfunction or disparities in sensor cover, or systematic failures in testing due to batch issues. These issues may be either transient, or represent unavoidable issues with the data capture process. For instance, data obtained from remote sensing equipment via satellite may be missing observations at certain times due to transient satellite malfunction, or be systematically limited for certain geographical regions due to restrictions in satellite orbits or fields of view (as occurs, for instance, at the so-called the `polar hole' \cite{andersson2021seasonal}).   

\emph{Skip patterns:} Fourth, SM commonly arises in survey data, in the form of `skip patterns' \cite{groves2011survey}. A skip pattern occurs when a set of responses are only obtained if the response to an earlier question is `Yes' with a `No' response leading to the participant skipping to a later part of the survey. In such cases, large blocks of missingness arise as an inherent and anticipated feature of the study design.

\emph{Population heterogeneity:} Fifth, SM arises when considering patterns of human behaviour or characteristics. For example, individuals naturally have different interests and desires, exhibit different patterns of activity, and have different physical attributes. Data that amalgamates these behaviours or attributes from a large population \cite{ledford2020facebook} will naturally contain a large amount of SM that reflects the inherent diversity of the underlying population, both in intrinsic characteristics and behaviour. 

While not exhaustive, these examples highlight some of the difficulties of dealing with SM: in some cases, SM is anticipated and may be appropriate; while in others it may relate to both avoidable or unavoidable data capture shortcomings and can profoundly inhibit subsequent learning. Understanding and dealing with SM is therefore not a straightforward process: a suite of tools, adapted to different circumstances, is needed, as well as processes by which the community can build on these tools as our understanding of SM evolves.  

\end{mdframed}

\section{Conceptualising structured missingness} \label{S1}

\textcolor{black}{In this section we review the standard literature on missing data, and discuss how existing taxonomies do not fully capture the complexities of SM. We also review some common routes to SM and contextualise the problem using a real-world health database example.}

Suppose we have a $n\times p$ data matrix $X$, that corresponds to $n$ units measured on $p$ variables. Now consider a corresponding $n\times p$ indicator matrix $R$, with entries either 0 or 1, such that $R_{ij} = 0$ indicates that $X_{ij}$ is missing, while $R_{ij} = 1$ indicates that $X_{ij}$ is observed. We can decompose $X$ into its observed ($X_{\textrm{obs}}$) and missing data parts ($X_{\textrm{mis}}$) as follows:
    $$X_{\textrm{obs}} = \{X_{ij} : R_{ij} = 1 \} \quad \mbox{and} \quad X_{\textrm{mis}} = \{ X_{ij} : R_{ij} = 0 \}.$$
Using this terminology we can define two important fundamental concepts in missing data: (1) missing data \emph{mechanisms} and (2) missing data \emph{patterns}.

Missing data \emph{mechanisms} refer to the process by which missing values arise in the data. In 1976 Rubin proposed a taxonomy of missingness mechanisms based on considering the conditional distribution $p(R \, | \, X, \phi)$, which has since been adopted as the standard for categorising missing data mechanisms \cite{rubin1976}. This taxonomy models the probability that values are missing as a function of the variables in the data, where $p(\cdot)$ denotes the general functional form of the relationship, and $\phi$ represents parameters in the model that specify the exact relationship between $R$ and $X$. Specifically, Rubin proposed the following three classes of missingness: (1) data missing completely at random (MCAR). In this case, missing readings are independent of both the observed and unobserved data and there are no systematic differences between units for which there are missing data and those that are complete. For data that are MCAR, complete case analysis (i.e. simply excluding from analysis those units for which data is missing) may weaken statistical power but does not introduce bias. Mathematically, for data that is MCAR, $p(R|X, \phi) = p(R| \phi)$. (2) Missing at Random (MAR). In this case, missing readings in one variable are dependent on observations of another. For instance, when completing a questionnaire on mental health concerns, men may be less forthcoming than women. Thus, assuming that gender data is complete, missingness in the response data will relate to an observed characteristic. For data that are MAR, complete case analysis may introduce bias because it may select for subsets of units that are not representative of the data as a whole. Mathematically, for data that is MAR, $p(R|X, \phi) = p(R| X_{obs}, \phi)$ (3). Missing Not at Random (MNAR). In this case, missing readings in one variable are dependent on missingness in others, or systematic factors outside of the scope of the experiment or data collection process. For data that are MNAR, complete case analysis may introduce bias; moreover, since the source of this bias is not apparent it cannot easily be dealt with in subsequent analysis. Mathematically, for data that is MNAR, $p(R|X, \phi) = p(R| X, \phi)$ (i.e. no simplifying relationship exists within this framework).

\begin{figure}
\centering
\includegraphics[width=\textwidth]{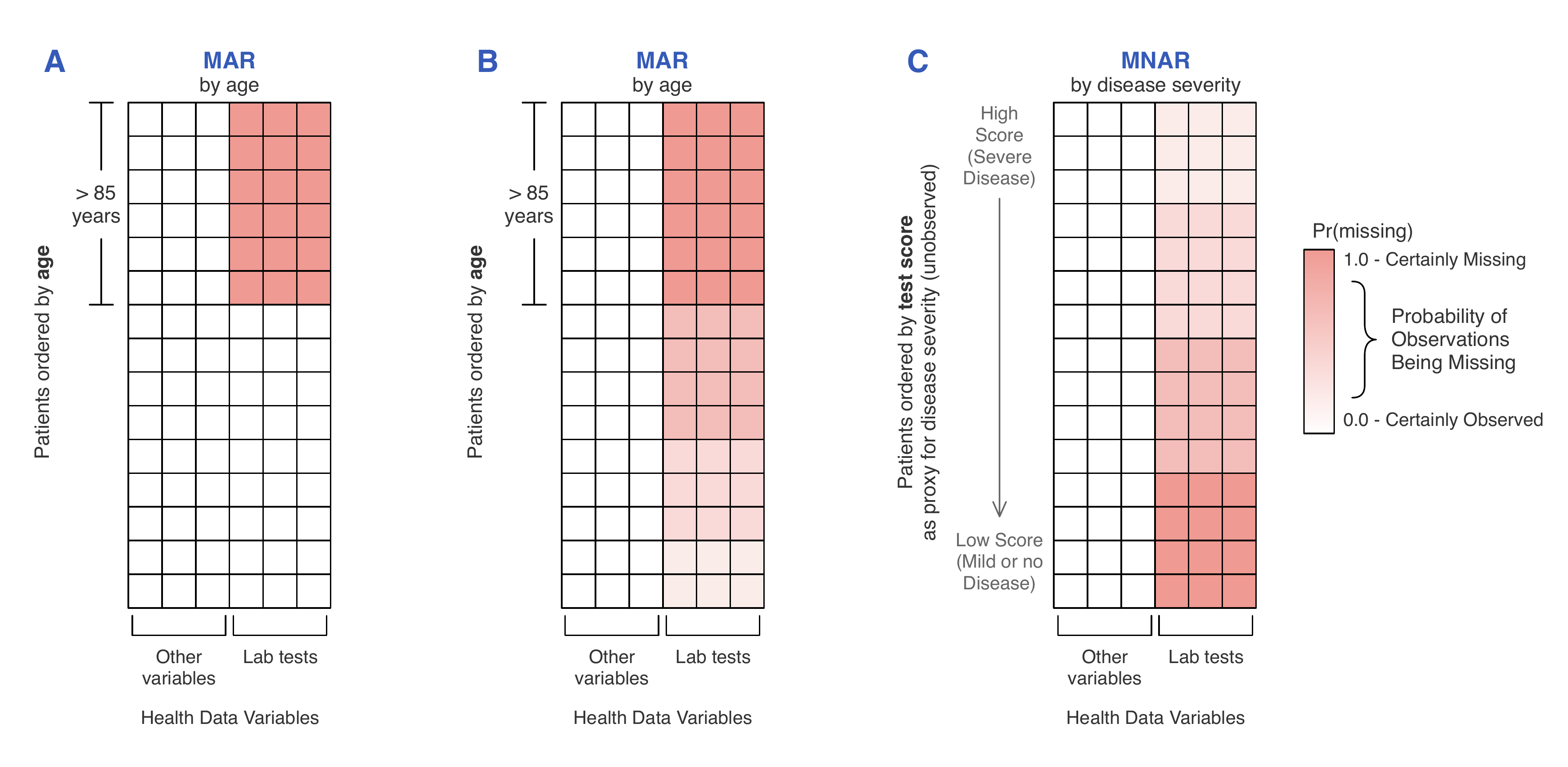}
\caption{\textbf{Examples of structured missingness}. Missingness can arise in subtle ways that are not well accounted for in classical theory. For example, a physician may or may not recommend a battery of tests to be conducted for a given patient depending on their health status or other pertinent characteristics such as age or sex, resulting in a set of variables in the data that are simultaneously missing or observed. Since clinician decisions are rarely binary, the resulting missingess relationships may be probabilistic rather than deterministic, resulting in a set of variables that are more likely to be missing jointly. (\textbf{A}) describes a hypothetical scenario in which lab tests are performed for all patients below a critical age (here, 85 years) but not for those above. This is an example of a monotone missing data pattern. (\textbf{B}) describes a more realistic scenario in which the probability of the test being performed decreases with increasing patient age. While a similar missingness mechanism (MAR), this gives rise to a non-monotone pattern. (\textbf{C}) describes a situation in which the physician is more likely to order a diagnostic lab test based suspected disease status that is revealed by the test itself (and so data is MNAR), which gives rise to a non-monotone pattern of SM. Here the SM pattern is shown for illustration only: in reality, the missingness pattern and structure would be unobservable as they depend on an unobserved variable -- the diagnostic test score which is a proxy for latent disease severity.}
\label{SMplot2}
\end{figure}

Missing data \emph{patterns} refer to the way in which missing values are located in the data. Although there are many possible ways to quantify patterns (see Challenge 2 below), in practice two categorisations are commonly used: monotone and non-monotone patterns (see \textbf{Fig}. \ref{SMplot2} and \cite{littlerubin2019}). A monotone pattern refers to a pattern where it is possible to order the $p$ variables in such a way that when $R_{ij} = 0$, $R_{ik} = 0$ for all $k > j$. A non-monotone pattern refers to a pattern where it is not possible to order the variables in this way. Monotone missing data patterns typically allow cleaner decomposition of the data into its missing ($X_{\textrm{mis}}$) and observed ($X_{\textrm{obs}}$) parts, which in turn can facilitate the use of methods that can obtain results analytically \cite{littlerubin2019}. However, in highly multivariate settings non-monotone patterns are much more likely to occur. Here, results can typically only be approximated using computational methods such as Markov Chain Monte Carlo techniques, although often these approximations can still be highly accurate. While missing data patterns may seem to share similarities with missing data mechanisms, the two are different and distinct characterisations of missing data. For example, non-monotone patterns can arise from MCAR, MAR, or MNAR mechanisms in the data, or indeed a mix of all three. While relationships between patterns and mechanisms are hard to specify in general, some simple rules of thumb do exist: for example, monotone patterns are unlikely to arise from MCAR mechanisms, due to the complex structured form that they typically take.

Rubin's missingness mechanisms and patterns are important because they provide a powerful guide to identifying different types of missingness; when issues such as bias are likely to arise and when not; and when tools, such as imputation, will be effective, and when not. For example, there are now numerous powerful methods for dealing with missing data assuming that it is MCAR, including a variety of imputation tools \cite{littlerubin2019}. Readings that are MAR are harder to address, but can sometimes be dealt with using more sophisticated imputation tools, such as multivariate imputation by chained equations (MICE) \cite{van2011mice}. Because they are the least precisely defined and most complex, readings that are MNAR are harder still to deal with, and are typically considered very cautiously, and dealt with in more pragmatic, bespoke ways: for example, via manual exploration and correction of missingness patterns, or by performing sensitivity analysis to determine how analysis results change under various data excision or imputation choices. 

However, despite their undoubted utility, Rubin's taxonomies do not fully account for the high-dimensional patterns of SM that are increasingly encountered in modern ML applications. For example, while it is clear that SM is associated with Rubin's MAR and MNAR categories, SM is not exclusively associated with either. An illustrative example is given in \textbf{Fig}. \ref{SMplot2}. Here, a physician decides which patients receive a battery of tests either based solely on their age (see \textbf{Fig}. \ref{SMplot2}A, B), or based on suspected disease status (see \textbf{Fig}. \ref{SMplot2}C). In cases where the decision is based on age, the data are MAR, because age is fully observed; whilst in the case where the decision is based on disease status, which is unknown, the data are MNAR and, importantly, the missing data structure is unobservable as it depends on the unknown variable -- a nefarious and silent consequence of MNAR. Moreover, depending on whether testing occurs in a deterministic or probabilistic fashion (i.e., if the clinician always refers patients of a given age for testing, as in \textbf{Fig}. \ref{SMplot2}A or uses their judgement based on other, perhaps unrecorded characteristics, as in \textbf{Fig}. \ref{SMplot2}B) the resulting patterns may be monotone or non-monotone. 

In high-dimensional settings -- in which numerous different generative mechanisms may overlay different patterns of missingness in this data -- such nuances can make it challenging to detect and characterise the different types of SM present in a dataset. For example, monotone patterns, such as block missing data patterns, often present in clear and obvious ways, and may be easy to visualise and characterise. On the other hand, non-monotone patterns, that arise from an underlying structure in which missing values are spread thinly across a highly multivariate space or in MNAR settings, can be much more pernicious and harder to characterise or even detect. 

\textcolor{black}{As appetite for learning from complex, high-dimensional data increases, such missing patterns will be increasingly encountered. To naively train an ML model on such data presents a danger because standard missing data methods do not sufficiently address the challenges of SM. For example, a complete-case analysis of the data in \textbf{Fig}. \ref{SMplot2}A (the default in many ML libraries) would remove all patients above 85 years of age, and thereby introduce bias by discarding an important section of the population. On the other hand, commonly-implemented imputation methods such as multiple imputation or tree-based imputation may either be inappropriate because the data are intentionally missing, or introduce high uncertainty depending on the number of variables and observations missing. Developing methods to decipher the (often complex) geometry of structured missingness is thus crucial in such settings (see Challenge 2, below).} 


In \textbf{Box} \ref{B1} we outline some common routes to SM; in \textbf{Box} \ref{B2} we provide a detailed motivating example of the issues that arise when dealing with SM in a complex real-world dataset. These examples illustrate that SM incorporates a very wide range of phenomena that do not fall neatly within current categories for missing data patterns or mechanisms, and each present their own challenges. In complex, multivariate settings these issues are further compounded: numerous different forms of SM may be present within a given dataset, each arising due to different mechanisms, each exhibiting its own patters, which may (or may not) relate to missingness patterns elsewhere in the data. This sheer diversity has hampered the formulation of SM as a field of study in its own right, and we consequently do not yet have the theoretical tools to classify the various common forms of SM, understand their implications for downstream learning or deal with them effectively.

\subsection*{Toward a taxonomy of structured missingness}
The examples highlighted above suggest that to learn from large, heterogeneous, linked datasets we need new ways of conceptualising SM and tools for dealing with it as an integrated part of the ML pipeline (see \textbf{Fig}. \ref{SMplot1}).  

There are several distinct mathematical and computational challenges associated with doing this. First, current taxonomies do not capture the richness of SM as phenomena and we lack the language to define and quantify both mechanisms and patterns of structured missingness: new mathematical formulations, that both extend and complement Rubin's classical taxonomy, and view SM as a fundamentally multivariate phenomenon, are needed. Indeed, simply detecting and categorising all the various types of SM present in a large multivariate dataset can be extremely challenging and there are, currently, no tools to do so to our knowledge. 

Second, lacking a formal framework to quantify SM, we lack the tools to assess the extent to which SM compromises or biases ML models. In some cases, missing data, even if significant in extent and highly structured, will not affect model performance. This may happen if, for instance, missing data are peripheral to the outcome of interest or when missing readings can be accurately inferred from observations. Yet, in other circumstances the presence of even modest SM can severely compromise model performance. This may happen if the missing data concerns a variable that is a causal determinant, or highly predictive, of an outcome of interest, for instance. While such issues may to some extent be addressed on a case-by-case basis, we currently lack a formal framework to address these issues in general in a principled way. 

Third, the impact that types of SM have on subsequent inference and learning has not yet been adequately explored. For example, if we assume a (perhaps unknown) relationship between variables and a subset of variables exhibits SM, what effect does this have on subsequent analysis involving other variables? 

Fourth, we have a very poor understanding of the effects that SM have on algorithmic bias and fairness. In highly sensitive and/or regulated areas, such as healthcare and criminal justice, in which ML tools are now beginning to be used \cite{leslie2021artificial, topol2019high}, these are not peripheral concerns: it is vital that any ML tools developed and deployed for public good address issues of fairness in a rigorous and transparent way \cite{macarthur_nature}. To do so, we need a better understanding of how SM affects algorithmic bias and the fairness of ML decisions.  

Based on these observations, the Alan Turing Institute convened a series of workshops to frame these issues and determine a trajectory for future research in SM. In the following section we outline nine grand challenges in SM, that arose from these discussions. These challenges may be grouped into the following four categories (illustrated in \textbf{Fig}. \ref{SMplot1}): (1) origins and definitions of SM; (2) measurements and modelling with SM; (3) inference and prediction with SM; (4) causality with SM. Collectively, they scope the problems associated with SM, provide a road map for the development of SM as a field of study, and outline a set of future research directions that will help advance ML at scale. 

\begin{mdframed}
\subsection*{Box 2: Structured missingness in practice: a motivating example} \label{B2}
The Flatiron Health-Foundation Medicine Clinico-Genomic Database (FH-FMI CGDB), henceforth simply referred to as CGDB, is a US nationwide, longitudinal, deidentified oncology database that combines real-world clinical data and outcomes for patients treated at approximately 280 US cancer clinics (approximately 800 sites of care) together with comprehensive genomic profiling of $\sim$600 unique cancer-related genes sequenced from over 400,000 samples. Retrospective longitudinal clinical data were derived from electronic health record (EHR) data, comprising patient-level structured and unstructured data, curated via technology-enabled abstraction, and were linked to genomic data derived from FMI comprehensive genomic profiling (CGP) tests in the CGDB by de-identified, deterministic matching \cite{singal2017cgdb,birnbaum2020modelassisted}. In total, the CGDB consists of data from $\sim$100,000 patients taken from a variety of disease-specific and disease-agnostic datasets.

Because of its breadth and complexity, the CGDB provides a compelling real-world example of the challenges of dealing with SM. Moreover, it illustrates how SM can arise in multiple different ways in the same dataset due to differences in data collection and linkage practices. To illustrate we discuss two: first, genomic samples in the CGDB are obtained from tests (known as assays) that each measure the alteration status of a specific set of genes. However, not all diagnostic assays are the same and assays often evolve over time. Patient A with a particular disease status may receive an assay comprising 400 genes, while Patient B of the same disease status may receive an assay comprising 150 genes, depending on the treatment or prognosis goals of the ordering physician. Any analysis of patients with the disease that combines records from Patient A and Patient B, and patients like them, must deal with the missing blocks of data for the genes not measured in both assays. 

Contrast this with a second SM data problem, in which certain data are \emph{not} clinically relevant (or possible) to measure for certain groups of patients. For example, prostate specific antigen (PSA) is a highly prognostic blood test used in screening for prostate cancer; similarly serum CA125 is widely used as a tumour marker for ovarian cancer. Combining specific cancer cohorts within the CGDB, such as patients with metastatic prostate cancer and ovarian cancer, will reveal patterns of SM in these cancer-specific variables (PSA will be unmeasured in female patients and CA125 in male patients). Other cancer-specific variables -- such as the Gleason Score, a prognostic grading score for patients with prostate cancer -- will also exhibit similar structures of missingness. Importantly, none would be appropriate to estimate outside of the relevant group of data collection (e.g. males/females, or prostate/ovarian cancer patients). However, broader analytic uses of the data -- for example, building prognostic models with cohorts containing males and females, or pan-cancer models constructed across multiple cancer types -- may merit the inclusion of these variables. In these cases, the occurrence of SM is not due to issues with data capture, but rather relates to the lack of clinical relevance of subsets of features to subsets of patients. Understanding and handling such patterns of SM poses a considerable methodological challenge when considering large, complex datasets that comprise information taken from numerous different sources, of which the CGDB is one example.
\end{mdframed}

\section{Structured missingness grand challenges} \label{S3}

\textcolor{black}{In this section we present nine grand challenges in addressing and mitigating the effects of SM. These challenges arose from a series of workshops held at the Alan Turing Institute in November-December 2021 and subsequent discussions.}

\subsection*{Challenge 1: Defining structured missingness}
Although missing data was a known problem before Rubin's seminal paper, its handling lacked scientific rigour and its causes were largely considered accidental and ignored \cite{rubin1976}. It was not until Rubin's formulation of missing data mechanisms that missingness gained a formal mathematical characterisation, from which arose strong active research areas devoted to the handling of missingness and more effective use of data with missing readings. We believe that the same is true of SM.

It is now clear that the concept of SM encompasses myriad kinds of missingness that are not fully described by Rubin's original taxonomy. Unarguably, it is difficult to address an ill-defined problem. Thus, to meaningfully handle SM, we must first rigorously define it. A key challenge is to identify the important types of SM, how they differ from existing formulations, such as MAR and MNAR, and how they relate to each other.

Three considerations are key. First, a defining feature of Rubin's taxonomy is its mathematical rigour: any systematic characterisation of SM must be similarly grounded in rigorous mathematical theory. A central challenge is developing a characterisation that meaningfully maps to existing mechanisms and patterns, and ensures that past developments can be embedded within this new context where appropriate. Second, any new definition must overcome the shortcomings of existing characterisations, including, for instance, overcoming certain ambiguities in interpreting MAR as initially proposed by Rubin \cite{seaman2013_2,doretti2018} and extending of MAR and MNAR assumptions to multivariate settings, which is neither straightforward nor intuitive \cite{tian2015missing}. Third, and perhaps most importantly, any formal characterisation for SM must be presented in a form that allows for its practical application: for instance, to determine when imputation is appropriate, and when not, and provide the foundation on which to build tools to do so as appropriate \cite{antelmi2021}.  

\subsection*{Challenge 2: Exploring the geometry of structured missingness}
SM is fundamentally characterised by non-random patterns of multivariate association between missing values. A simple yet potentially powerful way of characterising these patterns is through mathematical structures, such as networks and their generalisations, that capture their geometry and/or topology \textcolor{black}{and are thereby able to learn the underlying multivariate relationships between missing values. By so doing, the resulting networks can help address issues associated with gaps in the data, for instance by guiding the design of database enrichment strategies.} However, such geometric approaches to understanding missingness have not yet been extensively explored. 

Networks are a natural way to represent complex data and constitute a powerful alternative to linked tables. Network science can provide convenient tools to summarise structural properties of data, such as patterns of associations between variables \cite{newman2018networks} (the indicator matrix $R_{ij}$ can be considered as the incidence matrix of a network, for instance). However, network models, by definition, only capture pairwise interactions between variables. In a multivariate SM setting, higher-order interactions (i.e. coordinated missingness patterns in more than two variables) are of fundamental importance. Such higher-order associations may be captured by network generalisations such as simplicial complexes and hypergraphs \cite{Bianconi2021}, and decomposed using tools from information theory \cite{gutknecht2021bits}. Recent years have seen tremendous developments in the theory and practical applications of such higher order network theories and models, and related tools from geometry and topology, to data analytics \cite{bick2021higher,carlsson2009topology,joharinad2022geometry}. Importantly, these higher-order models are simultaneously amenable to statistical, topological and geometrical analysis, potentially allowing them to bring both statistical and geometric rigour to the study of SM. Similarly, networks can also be enriched by metadata -- for instance that classifies interactions according to different categories or layers -- giving rise to so-called multiplex networks \cite{Bianconi2018, kiani2021}. Collectively, these advances open up new tools for the study of SM. The challenge is to formulate novel computational and mathematical representations that, in combination with tools from information theory and the statistical sciences, are able to represent the architecture of SM in a statistically robust way and allow quantitative characterisation and visualisation of the geometrical and topological organisation of SM. 

\subsection*{Challenge 3: Design of experiments with structured missingness}
Because prevention is better than cure, SM should be considered at the data collection, or design, stage as much as possible. Accounting for missingness during the design phase of a study is already known to be advantageous \cite{Lee2019, Lee2018, Lee2018OptimalDF} and it is likely that this is particularly important in the more general setting of SM: for instance, to simplify any downstream analysis adversely affected by structural imbalances in the data. To address these challenges three things are particularly needed: (1) New tools for design of experiments to minimise the amount and impact of SM. While some SM may be inevitable, can the data be collected in a way that minimises its impact and anticipates future sources of SM? For example, in the future data from an experiment may be combined with data from other experiments it was not designed to be combined with. Can designs anticipate and account for such eventualities? To address this issue, the use of common standards should be explored. For example, development of a unifying framework for data collection and recording, with potential future variables in mind, would reduce the occurrence (and accumulation) of blocks of missingness. (2) Novel design methodology that facilitates characterisation and accounts for different forms of SM (see \textbf{Box} \ref{B1} and Challenge 1). Such methods could draw on and extend the theory of space filling designs \cite{Noonan2021, Zhigljavsky2020} to locate (and mitigate the effect of) particularly influential areas of missingness. (3) New tools that combine prior knowledge and domain expertise to anticipate forms of SM and incorporate them seamlessly into design methodology. If no prior knowledge of the form of SM is available, then methods that draw on theory of adaptive designs \cite{Burnett2020} could be considered. Two stage adaptive designs may prove particularly useful, in which the stage 1 design detects the forms of SM present in the data, and then the stage 2 design is optimised to mitigate their effects on downstream analysis. 

\subsection*{Challenge 4: Prediction with structured missingness}
Predictive modelling is a common objective for generating insights from multi-modal, heterogeneous data. However, in many predictive ML studies, missing data are poorly reported or accounted for, leading to compromised performance \cite{nijman2022missing}. Prediction from complex data with SM is a particular challenge because many ML algorithms that are equipped to handle complex multivariate relationships require complete data with no (or few) missing values. For example, while much progress has been made in the development of unsupervised deep generative models -- which are able to produce synthetic data with complex statistical associations that match those found in training data  --  relatively little attention has been given to dealing with missing data in supervised deep learning settings \cite{ipsen2020deal}. Making advances here requires developing the following five areas: (1) New tools to better understand the relationships between structures present in data and patterns of missingness and incorporate these patterns into predictive models (see also Challenge 2). (2) New imputation techniques adapted to predictive tasks that account for and, where appropriate, take advantage of the missingness structures (see Challenge 6). (3) New ways to combine models that learn from substructures within data and combine to produce holistic models that include both fully observed- and partially-observed features. (4) New tools that allow robust predictions of SM itself, including the ability to accurately predict patterns of missingness beyond training data. (5) New ways to assess model performance and uncertainty, accounting for missingness structures (see Challenge 7). For example, in a clinical context, model predictions may be more accurate for patients with genetic data than for patients who have not received genetic testing. Tools to assess differential performance for both categories of patients, and therefore a proxy assessment of the impact of missingness of genetic data on model performance, are required.

\subsection*{Challenge 5: Inference and estimation with structured missingness}
Statistical inference is the process of estimating properties of a population of interest from data. The presence of SM can significantly compromise this process. For example, if missing values are predominantly located in one part of the data (for instance, relating to a particular sub-population of individuals), then estimation procedures based on this data are likely to have higher levels of uncertainty than those based on more complete parts of the data. In ML contexts this can give rise to biased models that perform well on subsets of the data but poorly on others. For example, it has been recognised that a range of automated facial recognition algorithms can be highly accurate in identifying lighter-skinned male faces, but are uncertain in recognising darker-skinned female faces \cite{buolamwini2018gender}. This uncertainty is due, in part, to under-representation of darker-skinned subjects in underlying training datasets and the resulting inability of models to accurately learn features related to this group \cite{Leslie2020}. In this case, model uncertainty imparts useful information concerning missingess patterns that can and should be resolved. These issues therefore have profound implications for the fairness of resulting tools, and so are not just of technical importance (see Challenge 9).

Moreover, some forms of inference may not be possible under some types of SM. For example, if two categorical variables are simultaneously missing for a given level in each variable, then estimating the parameters for the interaction of those levels will not be possible under standard likelihood based inference. Bayesian inferential procedures may offer ways to mitigate some of these problems by imposing informative prior distributions (for instance, drawing on domain expert knowledge) which allow estimation of model parameters \cite{gelman2013}. Bayesian methods can also incorporate uncertainty, including model uncertainty, into inferences in a natural way, and allow complex models to be fitted to the data through computational methods such as Markov Chain Monte Carlo \cite{gelfand1990}. However, we currently lack rigorous ways to incorporate SM into such Bayesian frameworks and tools to assess when inference can be reliably performed and when not. Moreover, we also lack tools to make inference in the presence of different kinds of missingness. For instance, if missingness in one part of the data is MNAR and MAR in another, but the missing values also exhibit a general global structure (see Challenge 2), how are inferences affected at the local level, e.g. when restricted to data only containing MAR missingness? \textcolor{black}{Moreover, how do global patterns of SM affect the explainability of such inferences?} The key challenges are to develop inference tools that explicitly account for SM \textcolor{black}{without compromising explainability}, decouple the effects of different missingness mechanisms on subsequent inference, and are able to generate unbiased \textcolor{black}{and interpretable} estimates of population characteristics that allow conclusions to be drawn with confidence.

\subsection*{Challenge 6: Imputing structured missingness}
Imputation is the process of replacing missing (or dubious) values with plausible representative values based on the observed data \cite{van2018flexible}. Often imputations are performed multiply to obtain estimates of uncertainty or variability \cite{van2011mice}. The resulting ensembles can then either be used to estimate parameters for an implied generating model or imputed values can be used as direct input into subsequent analysis, such as formulating a predictive model. 


While there are numerous ways to impute data that is MCAR, imputation in the presence of SM presents both challenges and opportunities. When missing data arises in block structures, the wisdom of imputing a large distinct portion of the data in which a large fraction of data is missing is questionable, since it will likely lead to unreliable imputed values. In addition, for some types of SM, imputation may not be appropriate (see \textbf{Box} \ref{B2} for an example). Importantly, in complex, multivariate datasets that exhibit significant SM, being able to determine when imputation is appropriate and when it is not, may not be straightforward and could depend on numerous factors that might be intrinsic to the data, such as the imputed values of other missing data, or extrinsic, such as domain specific knowledge. Rigorous methodologies to address this problem, including incorporation of prior knowledge, are sorely needed.

In some circumstance SM also presents opportunities to build better informed imputation models that utilise the missing values themselves as information. For example, in a (hypothetical) study on sexual health, men might be reluctant to answer a question on loss of sexual function. If we subsequently observe a record in the study with missing responses on both gender and sexual function, we might be confident that the missing gender is likely to be male and we can build this information into our imputation model and/or any downstream learning. 

\subsection*{Challenge 7: Benchmarking and evaluation of structured missingness tools}
Common practice when assessing a new technique that deals with missing data follows two steps: the first is to artificially introduce missing readings with a predefined structure into an otherwise complete dataset (which could be either simulated or real). This process is called `amputation' \cite{Schouten2018, Brand1999, Brand2003}. The second is to perform the new technique (for instance, imputation) and any relevant subsequent analysis (such as classification or inference) and compare the results, according to an appropriate success metric (such as balanced accuracy, or related measures, for classification) with those obtained without amputation. 

In principle, evaluation of SM methods can follow this same process. However, in complex multivariate settings, amputating the data in a way that accurately reflects the SM patterns of interest can itself be a significant challenge. The standard approach of removing observations one variable at a time does not usually work, since it fails to capture intended multivariate missingness patterns \cite{Schouten2018}. Simulating complex multivariate missingess structures requires new tools to learn and quantify patterns of missingness themselves (see Challenges 2 and 4). Thus, the problem of benchmarking is inextricably entangled with the problem of understanding of SM: that is, to benchmark SM tools requires a prior understanding of SM, the very thing that the tools are designed to provide. Innovative new ways of benchmarking missingess are needed to avoid this impasse. 


\subsection*{Challenge 8: Causality with structured missingness}
It is well understood that missing data typically compromises causal inference. The presence of SM can do so particularly severely. For instance, if data is systematically missing for a given set of variables, then their causal relationship to an outcome of interest cannot be determined. Given its inherent difficulties, literature on causal inference with missing data is very scarce, and is particularly challenging when both input variables and potential outcomes are systematically missing \cite{mayer2021causal}. Thus, even if one can overcome the challenge of characterising the missingness structure of data (see Challenges 1-2), it is yet another challenge to measure the impact of different missingness structures on causal inference processes and develop benchmark tools to address these issues (see Challenge 7). 

There are two further reasons why causal inference is inextricably linked to SM. First, causal inference has an important role to play in ensuring algorithmic fairness or verification thereof \cite{Kusner2017} and so questions of causality are inherently related to issues of bias, inference and prediction (see Challenges 4, 5, and 9). For instance, biomedical datasets are often biased with respect to protected attributes like gender and race. In some cases, this is for genuine physiological reasons: heart disease datasets may contain more samples from male patients than female patients; skin cancer datasets may contain more patients with lighter skin than darker skin. While these biases may accurately represent patient demographics and hence be unbiased in a statistical sense, they naturally lead to class imbalances and are so are biased from a machine learning perspective, and can lead to algorithms that systematically fail when considering certain patient groups. Importantly, causal tools can help identify these biases and point toward ways to mitigate them \cite{Shen2021}. Developing new causal inference methods that are able to cope with SM -- or identify subsets of data that must be improved before causal questions can be addressed -- are vital to advancing ML at scale in a fair and transparent way. 

Second, causal inference provides a framework for thinking about SM holistically. Indeed, a causal inference problem can be re-framed as a missing data problem \cite{ding2017} such that for all possible outcomes, there is at most one potential outcome which can be observed, with the remaining ones unobserved or `missing'. The causal inference objective, then, is to estimate these inherently missing outcomes (i.e., counterfactuals) using appropriate methods. Extended to SM, there may be much to learn about how to estimate structures of missingness via causal inference methods such as inverse probability weighting and imputation \cite{seaman2013, sun2017, westreich2015}. A key challenge will be understanding and quantifying the impact of both measured and unmeasured values on the estimand of interest. 

\subsection*{Challenge 9: Ethical implications of structured missingness}
SM can arise from sociocultural biases at every stage of the data capture and analysis life-cycle (see \textbf{Fig}. \ref{SMplot1}), ranging from what data are captured and how, through to data processing, analysis, and use in decision-making \cite{verheij2018}. As a result, certain (sets of) variables may exhibit higher levels of missingness in particular sociodemographic subgroups of the population \cite{Kiang2021, Tsiampalis2020}. For example, individuals from protected groups may be reluctant to provide potentially damaging, or sensitive information. Healthcare practitioners may also harbour conscious or unconscious biases against patients of marginalised groups and make different diagnostic and treatment decisions based on these biases \cite{Leslie2021}. In some cases, data from historically under-served groups may be missing entirely or nearly so. Biomedical datasets, for example, are disproportionately made up of samples from Western individuals -- in genomics, for instance, 86\% of samples are from individuals of European descent \cite{Fatumo2022}, while in human microbiome studies, over 70\% of samples come from Europe, the United States, or Canada \cite{Abdill2022}. 

The extent to which SM is ethically problematic depends on the purpose of the study. Problems arise when biases result in a lack of generalisability to a wider population, or when the importance of explanatory factors is not understood because particular variables are missing or under-powered (see also Challenge 8). As a start, introducing standardised processes for documenting missingness in datasets would help ensure any conclusions appropriately acknowledge these biases \cite{Gebru2021, Rostamzadeh2022}. 

While in certain scenarios SM may be evident, more insidious settings can occur if it goes undetected. Methods to quantify SM according to certain sociodemographic characteristics may highlight the presence of societal biases in the data \cite{Tierneye007450}. Appropriate sensitivity to social processes that underlie data generation and contextual awareness of potential social, cultural, and historical determinants of discriminatory patterns are crucial for effective bias mitigation. Thus, involving experts with domain knowledge and social scientific training is vital. 


If not properly accounted for, SM may perpetuate or exacerbate existing inequalities \cite{Fernando2021, Martin2019}. While progress towards each of the challenges described above has the potential to mitigate such harms, addressing the problem fully will require explicit cross-disciplinary efforts directed towards issues of bias and fairness, connecting statistical questions of representativeness or homogeneity to pertinent ethical principles and values. \textcolor{black}{Thus, ethical considerations should be central to attempts to tackle all the SM challenges we have described. For instance, benchmarking and evaluation of SM tools should include considerations of fairness; methods for prediction and imputation should consider biases that may arise as a result of SM; and steps should be taken to ensure the collection of data where SM may occur is as representative as possible.}

\section{Conclusions} \label{S4}
Recent years have seen rapid growth in the availability of cheap digital storage and computational power, alongside disruptive advances in machine learning (ML) methods that are able to learn from vast quantities of data -- including data that are collected, curated and increasingly made available for research purposes -- and perform numerous downstream tasks \cite{bommasani2021opportunities}. Yet, these advances also come with their own challenges. While extraordinary progress has been made in specific ML areas such as natural language processing and computer vision, it is notable that these areas often rely on large freely available training data resources \cite{bansal2021}. In other important areas, such as healthcare and socio-economic policy, training data is not so easily obtained \cite{travers2018}, and is often compiled as an amalgamation of numerous separate resources, derived from disparate real-world experiments, that may not have been produced with amalgamation in mind. In such cases, the merged data required for model training is often fraught with missing or incomplete data\textcolor{black}{, alongside other well-recognized issues \cite{liang2022}}. These issues are particularly acute when dealing with `living' datasets that accrue new information with time \cite{koch2021}, merge it with old, and incorporate data from new measurement modalities -- such as evolving advances in genetic sequencing \cite{heather2016sequence} -- as they arise. For ML methods to learn from such dynamic, heterogeneous data, and generalise with robustness, they need to be designed to cope with the inevitable `structured missingness' (SM).  \textcolor{black}{These concerns are above and beyond issues of model degradation and bias associated with standard data cleaning processes \cite{li2021,krishnan2016}}. For this reason, we believe that there is now an urgent need to tackle SM as a topic in its own right, of central importance to the future of ML. To this end we have proposed a set of nine SM challenges that, if addressed, will boost our ability to learn effectively from real-world data at scale. \textcolor{black}{To address these challenges, two over-arching concerns are key.}

\textcolor{black}{First, a good understanding of the disciplinary or domain-context in which methods are employed is vital, as is an appreciation and clear articulation of their limitations. For instance, if computational or data resources do not permit a full investigation of the extent and impact of SM in an analysis pipeline, then clear communication of this fact, the potential biases that might arise and their practical implications should be acknowledged and unambiguously presented to stakeholders as part of the reporting process. Doing so will require a holistic approach to SM, within its wider context, involving multiple disciplines including ML, mathematics, network science, ethics and statistical design. There is thus an onus on hosting community-led cross-disciplinary workshops and development of platforms, such as open-source collaboration hubs, to best utilise the breadth of expertise to progress this emerging field.}

\textcolor{black}{Second, SM should not be considered only as a \emph{post hoc} problem, to be addressed at the analysis stage. To deal with SM most effectively it is important that its consideration enters at the design stage, and data collection strategies are designed to minimise the amount and effect of SM as far as possible and appropriate (see Challenge 3). Nevertheless, in some cases SM will be inevitable, or even appropriate (see Box 1), despite best efforts and, moreover, its presence may go undetected. Innovative new approaches, such as the use of network or topological models (see Challenge 2) or other data-driven approaches, such as unsupervised clustering, are needed, to facilitate better detection and characterisation of SM. Importantly, such \emph{post hoc} analysis tools are not independent of design considerations: SM will be most effectively tackled by a combination of good design and good analysis, as inter-related parts of an integrated analysis pipeline.}

To conclude, we offer some suggestions of possible routes to address the challenges we have proposed. While general solutions to these challenges will undoubtedly require the development of new mathematical and computational procedures, there are some existing tools and techniques that might help set promising trajectories for future research.

One possible way to tackle SM is to draw on the concepts of sparse and collaborative representation \cite{Zhang2011}. Collaborative representation models draw on the principles of collaborative filters, which are commonly used in recommender systems online \cite{Chakraborti2017}. Collaborative filters track the activities of many users over time -- each of which might have complex, specific patterns of missing data depending on their particular interests and profile characteristics (see \textbf{Box} \ref{B1}) -- and make personalised recommendations by comparing an individual's activity with patterns learnt from the community as a whole \cite{Schafer2007}. Collaborative representation classifiers \cite{Chakraborti2016} additionally use the fact that individual profiles can often be compactly represented in terms of a sparse basis set, which can be learnt by pooling data from the wider population. These approaches make use of two key facts that might help overcome some of the challenges of SM: (1) for some analyses, the effects of missing data can be mitigated by collecting and collating partial data in a distributed way over a large community. (2) In many cases, apparently high dimensional phenomena can be efficiently represented in a lower dimension, without significant loss of information, and this process of compression can be `denoising' (the human primary visual cortex uses such sparse encoding to effectively represent complex natural scenes \cite{vinje2000sparse}, for instance). 

A second possibility is to draw on recent concepts from synthetic data \cite{raghunathan2021synthetic, jordon2022synthetic}. Intuitively the problem of generating synthetic data can be considered as an extreme case of imputation in which \emph{all} the data is missing. As such, synthetic data generation tools may prove useful in imputing SM, where appropriate. Methods such as deep generative models, particularly self-attention transformers \cite{vaswani2017attention} and adversarial networks \cite{Zhang2018}, may prove particularly useful. For example, technologies used to generate deep fakes \cite{Mangaokar2021} and transfer style \cite{Xu2020} are beginning to be used in scientific applications, such as to normalise histopathology images \cite{Runz2021}. Similarly, powerful tools based these technologies have been developed to impute data that is MCAR with theoretical guarantees, and with empirical efficacy for data that is sparsely MAR and MNAR \cite{yoon2018gain}. While not yet able to deal with SM in all its forms, these are very promising developments. Moreover, they hint at the possibility of a virtuous circle in ML and SM research, in which advances in ML help us better understand SM, and advances in SM help us make the most of available data resources and develop more powerful ML methods.

\section*{Acknowledgments} \label{S5}
This work was sponsored by the Turing-Roche Strategic Partnership. The authors would like to thank Dr. Chloe Matus for her talents in figure illustrations and design and Vicky Hellon for her expert community management.

\section*{Competing interests}

The authors declare no competing interests.

\end{document}